\documentclass[journal,final]{IEEEtran}
\usepackage{graphicx,times,amsmath,amssymb,cite,subfigure,flushend,float}
\usepackage[lined,ruled,boxed]{algorithm2e}
\DeclareMathOperator*{\argmin}{\arg\!\min}
\IEEEoverridecommandlockouts \allowdisplaybreaks
\newtheorem{Definition}{Definition} 

\begin{document}

\title{Switching EEG Headsets Made Easy: \\ Reducing Offline Calibration Effort \\ Using Active Weighted Adaptation Regularization}

\author{\IEEEauthorblockN{Dongrui Wu, \textit{Senior Member, IEEE}, Vernon J. Lawhern, \textit{Member, IEEE}, \\ W. David Hairston, Brent J. Lance, \textit{Senior Member, IEEE}}\\
\thanks{Manuscript received June 16, 2015; revised January 13, 2016; accepted March 10, 2016. Research was sponsored by the U.S. Army Research Laboratory and was accomplished under Cooperative Agreement Numbers W911NF-10-2-0022 and W911NF-10-D-0002/TO 0023. The views and the conclusions contained in this document are those of the authors and should not be interpreted as representing the official policies, either expressed or implied, of the U.S. Army Research Laboratory or the U.S Government.}
\thanks{D. Wu is with DataNova, Clifton Park, NY USA (email: drwu09@gmail.com).}
\thanks{V. J. Lawhern is with the Human Research and Engineering Directorate, U.S. Army Research Laboratory, Aberdeen Proving Ground, MD USA. He is also with the Department of Computer Science, University of Texas at San Antonio, San Antonio, TX USA (emails: vernon.j.lawhern.civ@mail.mil).}
\thanks{W. D. Hairston and B. J. Lance are with the Human Research and Engineering Directorate, U.S. Army Research Laboratory, Aberdeen Proving Ground, MD USA (emails: william.d.hairston4.civ@mail.mil, brent.j.lance.civ@mail.mil). }
\thanks{Color versions of one or more of the figures in this paper are available online at http://ieeexplore.ieee.org.}
\thanks{Digital Object Identifier XXXXXXXXXXX.} } \maketitle

\begin{abstract}
Electroencephalography (EEG) headsets are the most commonly used sensing devices for Brain-Computer Interface. In real-world applications, there are advantages to extrapolating data from one user session to another. However, these advantages are limited if the data arise from different hardware systems, which often vary between application spaces. Currently, this creates a need to recalibrate classifiers, which negatively affects people's interest in using such systems. In this paper, we employ active weighted adaptation regularization (AwAR), which integrates weighted adaptation regularization (wAR) and active learning, to expedite the calibration process. wAR makes use of labeled data from the previous headset and handles class-imbalance, and active learning selects the most informative samples from the  new headset  to label.  Experiments on single-trial  event-related potential  classification  show  that AwAR can significantly increase the classification accuracy, given the  same  number of labeled  samples from the  new headset. In other words, AwAR can effectively reduce the number of labeled samples required from the new headset, given a desired classification accuracy, suggesting value in collating data for use in wide scale transfer-learning applications.
\end{abstract}


\section{Introduction}

\IEEEPARstart{E}{lectroencephalography} (EEG) headsets are the most commonly used sensing devices for Brain-Computer Interface (BCI), which have been employed in many applications, such as healthcare and gaming  \cite{Hamadicharef2010,Lance2012,Erp2012,Wolpaw2012,McDowell2013}, because of the general ease of setup for normal individuals. However, BCI applications have not received widespread acceptance for real-world applications. One reason for this is the inability of BCI technologies to adapt to the numerous potential sources of variation inherent in the underlying technologies. These can include human sources of variability, such as individual differences and intra individual variability. They can also include sources of variability in the technology, such as unintentional differences in recording locations for the  EEG electrodes from session to session, or even differences between different EEG headsets. To date, this latter source remains largely unexplored.

There are many existing EEG headsets, with new models and styles continually becoming available \cite{Hairston2014}. Ideally, EEG classification methods should be completely independent from any specific EEG hardware, such that classifiers trained using data from one EEG headset will be transferable to other headsets with little or no recalibration. This would help ensure that applications could reach a broad base of users and would not become obsolete through hardware upgrades. However, evidence comparing the performance of various classifiers when using different headsets has shown that often performance is not equal across systems; that is, the headset does in fact matter \cite{Ries2014}. From a hardware standpoint, systems can vary along a number of dimensions, including (but not limited to) onboard filter characteristics, electrode types and contact methods, electrode locations, or online reference schemes. All of these inherently change the resulting signal characteristics, some of which are critical features on which the classifiers operate.

Thus, it is not surprising that currently switching to a new or different headset requires the subject to re-calibrate it, which can take anywhere from 5-20 minutes \cite{Erp2012}. When implemented into a BCI system this calibration session would decrease the utility and appeal of the overall system, likely slowing the rate of acceptance. While it is not currently possible to switch between EEG headsets completely calibration-free, it is certainly possible to decrease the amount of time and data needed to calibrate an EEG data classifier for use with another EEG system.

In this paper, we specifically attempt to address the problem of developing classifiers that can account for variation due to different EEG headsets within a transfer learning (TL) \cite{Pan2010} framework. In TL, some data from a prior calibration or other user sessions is used to facilitate learning of the calibration in a new target context. According to a recent literature review \cite{Wang2015}, there are mainly three types of TL approaches for BCI applications:
\begin{enumerate}
\item \textit{Feature representation transfer} \cite{Satti2010,Kang2009,Lotte2010,Devlaminck2011,Samek2013,Spuler2012,Lee2009}, which encodes the knowledge across different subjects or sessions as features. These features are generally better than extracting features directly from only the limited number of samples from a new subject or session.
\item \textit{Instance transfer} \cite{Li2010,Li2009,drwuTL2011,drwuPLOS2013}, which uses certain parts of the data from other subjects or sessions to help the learning for the current subject or session. The underlying assumption is that data distributions for these subjects or sessions are similar.
\item \textit{Classifier transfer}, which includes domain adaptation \cite{Vidaurre2011,Bamdadian2013,Spuler2012}, i.e., handling the different data distributions for different subjects or sessions, and ensemble learning \cite{Tu2012,Tu2012b}, i.e., combining multiple classifiers from multiple subjects or sessions, and their combinations \cite{drwuSMC2015,drwuACII2015,drwuaBCI2015}.
\end{enumerate}
In our case, data acquired from one style of headset is used to facilitate classification of data currently being acquired from a different one, through domain adaptation and regularized optimization \cite{Tomioka2010,Zhang2013,Suykens2014}. We look at this problem within the context of offline single-trial Event-Related Potential (ERP) classification, with the eventual goal of moving to online single-trial classification within a BCI system.

In some application domains, we have existing unlabeled data and the calibration session is focused on labeling this data, e.g., BCI applications focused on labeling images, using EEG data  \cite{Tan2010,Uscumlic2013}. In these applications, the user  can  manually label  a  few  images,  and  based  on  the EEG signals associated with these images a classifier can be trained to automatically label the rest. Improved calibration performance can be achieved by selecting the most informative images for manual labeling. In other words, a desired level of calibration performance can be obtained with less labeling effort if the most informative images are selected for labeling. This is the idea of active learning (AL)  \cite{Settles2009}, which has also started to find application in BCI \cite{Chen2015,drwuSMC2015AL,drwuRSVP2016}. For example, in our recent work on EEG artifacts classification \cite{drwuSMC2015AL}, we showed that classification accuracy equivalent to classifiers trained on full data annotation can be obtained while labeling less than 25\% of the data by AL. In another study \cite{drwuRSVP2016}, we applied AL to a simulated BCI system for target identification using data from a rapid serial visual presentation paradigm, and showed that it can produce similar overall classification accuracy with significantly less labeled data (in some cases less than 20\%) when compared to alternative calibration approaches.

TL and AL are complementary to each other, and hence can be integrated to further reduce the number of labeled training samples in offline BCI calibration. The idea of integrating TL and AL was proposed recently  \cite{Shi2008} and is beginning to be explored \cite{Zhao2013,Chattopadhyay2013,Rai2010,Chen2011,drwuSMC2014}. However, most of this work is outside of the EEG analysis domain. In our previous work \cite{drwuSMC2014}, we investigated how TL and AL can be integrated to reduce the amount of subject-specific calibration data in a Visual-Evoked Potential (VEP) task, by making use of data collected using the same headset but from other subjects; in contrast, this paper considers the problem of reducing subject-specific calibration data when the same subject switches from one headset to another.

This paper introduces weighted adaptation regularization (wAR), a particular TL algorithm, and designs a novel AL algorithm for it. Using a single-trial ERP experiment, we demonstrate that wAR can achieve improved performance over the TL approach used in \cite{drwuSMC2014}, and active weighted adaptation regularization (AwAR), which integrates wAR and AL, can further reduce the offline calibration effort when switching between different EEG headsets. It should be noted that, while the ultimate goal is an understanding of how well these approaches work when transferring both within and across subjects, here, in order to minimize sources of variability, our analyses are focused on within subjects TL.

The rest of the paper is organized as follows: Section~\ref{sect:wAR} introduces the details of wAR. Section~\ref{sect:AwAR} introduces the details of AwAR. Section~\ref{sect:experiments} describes experimental results and a performance comparison of wAR and AwAR with other algorithms. Section~\ref{sect:conclusions} draws conclusions.

\section{Weighted Adaptation Regularization (wAR)} \label{sect:wAR}

This section introduces the details of the wAR algorithms. We consider two-class classification of EEG data, but the algorithms can also be generalizable to other calibration problems.

\subsection{Problem Definition}

Given a large amount of labeled EEG epochs from one headset, how can that data be used to customize a classifier for a different headset? Although EEG epochs from the two headsets are usually not completely consistent, previous data still contain useful information, due to the fact that they came from the same subject. As a result, the amount of calibration data may be reduced if these auxiliary EEG epochs are used properly.

TL \cite{Pan2010,Wu2004}, particularly wAR, is a framework for addressing the aforementioned problem. Some notations used in TL and wAR are introduced next.

\begin{Definition} \textbf{(Domain)} \cite{Pan2010,Long2014} A domain $\mathcal{D}$ is composed of a $d$-dimensional feature space $\mathcal{X}$ and a marginal probability distribution $P(\mathbf{x})$, i.e., $\mathcal{D}=\{\mathcal{X},P(\mathbf{x})\}$, where $\mathbf{x}\in \mathcal{X}$.
\end{Definition}

If two domains $\mathcal{D}_s$ and $\mathcal{D}_t$ are different, then they may have different feature space, i.e., $\mathcal{X}_s\neq \mathcal{X}_t$, and/or different marginal probability distributions, i.e., $P_s(\mathbf{x})\neq P_t(\mathbf{x})$ \cite{Long2014}.

\begin{Definition} \textbf{(Task)} \cite{Pan2010,Long2014} Given a domain $\mathcal{D}$, a task $\mathcal{T}$ is composed of a label space $\mathcal{Y}$ and a prediction function $f(\mathbf{x})$, i.e., $\mathcal{T}=\{\mathcal{Y},f(\mathbf{x})\}$.
\end{Definition}

Let $y\in \mathcal{Y}$, then $f(\mathbf{x})=Q(y|\mathbf{x})$ can be interpreted as the conditional probability distribution. If two tasks $\mathcal{T}_s$ and $\mathcal{T}_t$ are different, then they may have different label spaces, i.e., $\mathcal{Y}_s\neq \mathcal{Y}_t$, and/or different conditional probability distributions, i.e., $Q_s(y|\mathbf{x})\neq Q_t(y|\mathbf{x})$ \cite{Long2014}.

\begin{Definition} \textbf{(Domain Adaptation}) Given a source domain $\mathcal{D}_S=\{(\mathbf{x}_1,y_1),...,(\mathbf{x}_n,y_n)\}$, and a target domain $\mathcal{D}_T$ with $m_l$ labeled samples $\{(\mathbf{x}_{n+1},y_{n+1}),...,(\mathbf{x}_{n+m_l},y_{n+m_l})\}$ and $m_u$ unlabeled samples $\{\mathbf{x}_{n+m_l+1}, ..., \mathbf{x}_{n+m_l+m_u}\}$, domain adaptation transfer learning aims to learn a target prediction function $f: \mathbf{x}_t \mapsto y_t$ with low expected error on $\mathcal{D}_t$, under the assumptions $\mathcal{X}_s=\mathcal{X}_t$, $\mathcal{Y}_s=\mathcal{Y}_t$, $P_s(\mathbf{x})\neq P_t(\mathbf{x})$, and $Q_s(y|\mathbf{x})\neq Q_t(y|\mathbf{x})$.
\end{Definition}

In our application, EEG epochs from the new headset are in the target domain, while EEG epochs from the previous headset are in the source domain. A single data sample would consist of the feature vector for a single EEG epoch from a headset, collected as a response to a specific stimulus. Though the features in source and target domains are computed in the same way, generally their marginal and conditional probability distributions are different, i.e., $P_s(\mathbf{x})\neq P_t(\mathbf{x})$ and $Q_s(y|\mathbf{x})\neq Q_t(y|\mathbf{x})$, because the two headsets may have different sensor locations, filters, and signal fidelity. As a result, the auxiliary data from the source domain cannot represent the primary data in the target domain accurately and must be integrated with some labeled data in the target domain to induce the target predictive function.

\subsection{The Learning Framework}

Because
\begin{align}
f(\mathbf{x})=Q(y|\mathbf{x})=\frac{P(\mathbf{x},y)}{P(\mathbf{x})}=\frac{Q(\mathbf{x}|y)P(y)}{P(\mathbf{x})},
\end{align}
to use the source domain data in the target domain, we need to make sure\footnote{Strictly speaking, we should make sure $P_s(y)$ is also close to $P_t(y)$. However, in this paper we assume all subjects conduct similar VEP tasks, so $P_s(y)$ and $P_t(y)$ are intrinsically close. Our future research will consider the more general case that $P_s(y)$ and $P_t(y)$ are different.} $P_s(\mathbf{x}_s)$ is close to $P_t(\mathbf{x}_t)$, and $Q_s(\mathbf{x}_s|y_s)$ is also close to $Q_t(\mathbf{x}_t|y_t)$.

Let the classifier be $f=\mathbf{w}^T\phi(\mathbf{x})$, where $\mathbf{w}$ is the classifier parameters, and $\phi:\mathcal{X}\mapsto \mathcal{H}$ is the feature mapping function that projects the original feature vector to a Hilbert space $\mathcal{H}$. The learning framework of wAR is formulated as:
\begin{align}
f=&\argmin\limits_{f\in\mathcal{H}_K}\sum_{i=1}^{n}w_{s,i}\ell(f(\mathbf{x}_i),y_i)+w_t\sum_{i=n+1}^{n+m_l}w_{t,i}\ell(f(\mathbf{x}_i),y_i) \nonumber\\
& +\sigma\|f\|_K^2+\lambda_P D_{f,K}(P_s,P_t)+\lambda_Q D_{f,K}(Q_s,Q_t) \label{eq:f}
\end{align}
where $\ell$ is the loss function, $w_t$ is the overall weight of target domain samples, $K\in R^{(n+m_l+m_u)\times(n+m_l+m_u)}$ is the kernel function induced by $\phi$ such that $K(\mathbf{x}_i,\mathbf{x}_j)=\langle\phi(\mathbf{x}_i),\phi(\mathbf{x}_j)\rangle$, and $\sigma$, $\lambda_P$ and $\lambda_Q$ are non-negative regularization parameters. $w_t$ is the overall weight for target domain samples, which should be larger than 1 so that more emphasis is given to target domain samples than source domain samples. $w_{s,i}$ is the weight for the $i^\mathrm{th}$ sample in the source domain, and $w_{t,i}$ is the weight for the $i^\mathrm{th}$ sample in the target domain, i.e.,
\begin{align}
w_{s,i}&=\left\{\begin{array}{ll}
                  1, & \mathbf{x}_i\in \mathcal{D}_{s,1} \\
                  n_1/(n-n_1), & \mathbf{x}_i\in \mathcal{D}_{s,2}
                \end{array}\right. \label{eq:ws}\\
w_{t,i}&=\left\{\begin{array}{ll}
                  1, & \mathbf{x}_i\in \mathcal{D}_{t,1} \\
                  m_1/(m_l-m_1), & \mathbf{x}_i\in \mathcal{D}_{t,2}
                \end{array}\right.  \label{eq:wt}
\end{align}
in which $\mathcal{D}_{s,c}=\{\mathbf{x}_i|\mathbf{x}_i\in \mathcal{D}_s\wedge y_i=c\}$ is the set of samples in Class $c$ of the source domain, and $\mathcal{D}_{t,c}=\{\mathbf{x}_j|\mathbf{x}_j\in \mathcal{D}_t\wedge y_j=c\}$ is the set of samples in Class $c$ of the target domain, $n_c=|\mathcal{D}_{s,c}|$ and $m_c=|\mathcal{D}_{t,c}|$. The goal of $w_{s,i}$ and $w_{t,i}$ is to balance the number of positive and negative samples in source and target domains, respectively.

Briefly speaking, the meanings of the five terms in (\ref{eq:f}) are:
\begin{enumerate}
\item The 1st term minimizes the loss on fitting the labeled samples in the source domain.
\item The 2nd term minimizes the loss on fitting the labeled samples in the target domain.
\item The 3rd term minimizes the structural risk of the classifier.
\item The 4th term minimizes the distance between the marginal probability distributions $P_s(\mathbf{x}_s)$ and $P_t(\mathbf{x}_t)$.
\item The 5th term minimizes the distance between the conditional probability distributions $Q_s(\mathbf{x}_s|y_s)$ and $Q_t(\mathbf{x}_t|y_t)$.
\end{enumerate}
By the Representer Theorem \cite{Belkin2006,Long2014}, the solution of (\ref{eq:f}) admits an expression:
\begin{align}
f(\mathbf{x})=\sum_{i=1}^{n+m_l+m_u}\alpha_iK(\mathbf{x}_i,\mathbf{x})=\boldsymbol{\alpha}^TK(X,\mathbf{x}) \label{eq:f2}
\end{align}
where $X=[\mathbf{x}_1, ...,\mathbf{x}_{n+m_l+m_u}]^T$, and $\boldsymbol{\alpha}=[\alpha_1,...,\alpha_{n+m_l+m_u}]^T$ are coefficients to be computed.

Note that our algorithm formulation and derivation closely resemble those in \cite{Long2014}; however, there are several major differences:
\begin{enumerate}
\item We consider the scenario that there are a few labeled samples in the target domain, whereas \cite{Long2014} assumes there are no labeled samples in the target domain.
\item We explicitly consider the class imbalance problem in both domains by introducing the weights on samples from different classes.
\item wAR is iterative and we further design an AL algorithm for it, whereas in \cite{Long2014} domain adaptation is performed only once and there is no AL.
\item \cite{Long2014} also considers manifold regularization \cite{Belkin2006}. We investigated it, but we were not able to achieve improved performance in our application, so we excluded it in this paper.
\end{enumerate}
Also note that one of the wAR algorithms (wAR-RLS) described in this paper was introduced in our previous publication \cite{drwuSMC2015}; however, this paper includes a new wAR algorithm (wAR-SVM), and shows how AL can be integrated with wAR-RLS and wAR-SVM. The application scenario is also different.

\subsection{Loss Functions Minimization}

Two widely used loss functions are the squared loss for regularized least squares (RLS):
\begin{align}
\ell(f(\mathbf{x}_i),y_i)=(y_i-f(\mathbf{x}_i))^2 \label{eq:l2}
\end{align}
and the hinge loss for support vector machines (SVMs):
\begin{align}
\ell(f(\mathbf{x}_i),y_i)=\max(0,1-y_if(\mathbf{x}_i)) \label{eq:l1}
\end{align}
Both will be considered in this paper. In the following, we denote the classifier obtained using squared loss as wAR-RLS, and the one obtained using hinge loss as wAR-SVM.

\subsubsection{Squared Loss}

Let
\begin{align}
\mathbf{y}=[y_1,...,y_{n+m_l+m_u}]^T \label{eq:y}
\end{align}
where $\{y_1,...,y_n\}$ are known labels in the source domain, $\{y_{n+1},...,y_{n+m_l}\}$ are known labels in the target domain, and $\{y_{n+m_l+1},...,y_{n+m_l+m_u}\}$ are pseudo labels for the unlabeled target domain samples, i.e. labels estimated using another classifier and known samples in both source and target domains.

Define $E\in R^{(n+m_l+m_u)\times(n+m_l+m_u)}$ as a diagonal matrix with
\begin{align}
E_{ii}=\left\{\begin{array}{ll}
                w_{s,i}, & 1\le i\le n \\
                w_tw_{t,i}, &  n+1 \le i \le n+m_l\\
                0, & \text{otherwise}
              \end{array}\right. \label{eq:E}
\end{align}

Substituting (\ref{eq:l2}) into the first two terms in (\ref{eq:f}), it follows that
\begin{align}
&\sum_{i=1}^{n}w_{s,i}\ell(f(\mathbf{x}_i),y_i)+w_t\sum_{i=n+1}^{n+m_l}w_{t,i}\ell(f(\mathbf{x}_i),y_i) \nonumber \\
=&\sum_{i=1}^{n}w_{s,i}(y_i-f(\mathbf{x}_i))^2+w_t\sum_{i=n+1}^{n+m_l}w_{t,i}(y_i-f(\mathbf{x}_i))^2 \nonumber \\
=&\sum_{i=1}^{n+m_l+m_u}E_{ii}(y_i-f(\mathbf{x}_i))^2 \nonumber \\
=&(\mathbf{y}^T-\boldsymbol{\alpha}^TK)E(\mathbf{y}-K\boldsymbol{\alpha}) \label{eq:l3}
\end{align}

\subsubsection{Hinge Loss}

Using the hinge loss and $E$ defined in (\ref{eq:E}), the first two terms on the right-hand side of (\ref{eq:f}) can be re-expressed as:
\begin{align}
&\sum_{i=1}^{n}w_{s,i}\ell(f(\mathbf{x}_i),y_i)+w_t\sum_{i=n+1}^{n+m_l}w_{t,i}\ell(f(\mathbf{x}_i),y_i) \nonumber \\
=&\sum_{i=1}^{n}w_{s,i}\max(0,1-y_if(\mathbf{x}_i))\nonumber\\
&\quad +w_t\sum_{i=n+1}^{n+m_l}w_{t,i}\max(0,1-y_if(\mathbf{x}_i)) \nonumber \\
=&\sum_{i=1}^{n+m_l+m_u}E_{ii}\max\left(0,1-y_if(\mathbf{x}_i)\right) \label{eq:HingeLoss}
\end{align}

Often in SVM formulations, an unregularized bias term $b$ is added to (\ref{eq:f2}), i.e.,
\begin{align}
f(\mathbf{x})=\sum_{i=1}^{n+m_l+m_u}\alpha_iK(\mathbf{x}_i,\mathbf{x})+b=\boldsymbol{\alpha}^TK(X,\mathbf{x})+b \label{eq:f3}
\end{align}
We also use this convention in this paper. Then, by introducing non-negative slack variables $\xi_i$ ($i=1,2,...,n+m_l+m_u$), the minimization of (\ref{eq:HingeLoss}) is equivalent to:
\begin{align}
&\min\limits_{\boldsymbol{\alpha}\in R^{n+m_l+m_u}\atop \boldsymbol{\xi}\in R^{n+m_l}}\sum_{i=1}^{n+m_l}E_{ii}\xi_i \label{eq:l4}\\
&\text{s.t.}\quad y_i\left[\sum_{j=1}^{n+m_l+m_u}\alpha_j K(\mathbf{x}_i,\mathbf{x}_j)+b\right]\ge 1-\xi_i\nonumber\\
&\qquad \xi_i\ge 0,\qquad i=1,...,n+m_l \nonumber
\end{align}

\subsection{Structural Risk Minimization}

As in \cite{Long2014,Vapnik1998}, we define the structural risk as the squared norm of $f$ in $\mathcal{H}_K$, i.e.,
\begin{align}
\|f\|_K^2&=\sum_{i=1}^{n+m_l+m_u} \sum_{j=1}^{n+m_l+m_u}\alpha_i\alpha_jK(\mathbf{x}_i,\mathbf{x}_j)=\boldsymbol{\alpha}^T K\boldsymbol{\alpha} \label{eq:fK}
\end{align}

\subsection{Marginal Probability Distribution Adaptation}

Similar to \cite{Long2014,Quanz2009}, we compute $D_{f,K}(P_s,P_t)$ using the projected maximum mean discrepancy (MMD):
\begin{align}
D_{f,K}(P_s,P_t)=&\left[\frac{1}{n}\sum_{i=1}^nf(\mathbf{x}_i)-\frac{1}{m_l+m_u}\sum_{i=n+1}^{n+m_l+m_u}f(\mathbf{x}_i)\right]^2 \nonumber \\ =&\boldsymbol{\alpha}^TKM_0K\boldsymbol{\alpha} \label{eq:DfKP}
\end{align}
where $M_0\in R^{(n+m_l+m_u)\times (n+m_l+m_u)}$ is the MMD matrix:
\begin{align}
(M_0)_{ij}=\left\{\begin{array}{ll}
                             \frac{1}{n^2},& 1\le i \le n, 1\le j \le n  \\
                             \frac{1}{(m_l+m_u)^2}, & n+1\le i \le n+m_l+m_u, \\
                             & n+1\le j \le n+m_l+m_u \\
                             \frac{-1}{n(m_l+m_u)}, & \text{otherwise}
                           \end{array}\right. \label{eq:M0}
\end{align}

\subsection{Conditional Probability Distribution Adaptation}

Similar to the idea proposed in \cite{Long2014}, we first need to compute pseudo labels for the unlabeled target domain samples and construct the label vector $\mathbf{y}$ in (\ref{eq:y}). These pseudo labels can be borrowed directly from the estimates in the previous iteration if the algorithm is used iteratively, or estimated using another classifier, e.g., a SVM. We then compute the projected MMD w.r.t. each class. The distance between the conditional probability distributions in source and target domains is next computed as:
\begin{align}
&D_{f,K}(Q_s,Q_t)\nonumber \\
=&\sum_{c=1}^2\left[\frac{1}{n_c}\sum\limits_{\mathbf{x}_i\in \mathcal{D}_{s,c}} f(\mathbf{x}_i)-\frac{1}{m_c}\sum\limits_{\mathbf{x}_j\in\mathcal{D}_{t,c}}f(\mathbf{x}_j)\right]^2 \label{eq:DfKQ}
\end{align}
where $\mathcal{D}_{s,c}$, $\mathcal{D}_{t,c}$, $n_c$ and $m_c$ have been defined under (\ref{eq:wt}).

Substituting (\ref{eq:f2}) into (\ref{eq:DfKQ}), it follows that
\begin{align}
&D_{f,K}(Q_s,Q_t)\nonumber \\
=&\sum_{c=1}^2\left[\frac{1}{n_c}\sum\limits_{\mathbf{x}_i\in \mathcal{D}_{s,c}} \boldsymbol{\alpha}^TK(X,\mathbf{x}) -\frac{1}{m_c}\sum\limits_{\mathbf{x}_j\in\mathcal{D}_{t,c}}\boldsymbol{\alpha}^TK(X,\mathbf{x})\right]^2 \nonumber\\
=&\sum_{c=1}^2\boldsymbol{\alpha}^TKM_cK\boldsymbol{\alpha} =\boldsymbol{\alpha}^TKMK\boldsymbol{\alpha} \label{eq:DfKQ2}
\end{align}
where
\begin{align}
M=M_1+M_2 \label{eq:M}
\end{align}
in which $M_1$ and $M_2$ are MMD matrices computed as:
\begin{align}
(M_c)_{ij}=\left\{\begin{array}{ll}
                    1/n_c^2, & \mathbf{x}_i, \mathbf{x}_j\in \mathcal{D}_{s,c} \\
                    1/m_c^2, & \mathbf{x}_i, \mathbf{x}_j\in \mathcal{D}_{t,c} \\
                    -1/(n_cm_c), & \mathbf{x}_i\in\mathcal{D}_{s,c}, \mathbf{x}_j\in \mathcal{D}_{t,c}, \text{ or }\\
                    & \mathbf{x}_j\in\mathcal{D}_{s,c}, \mathbf{x}_i\in \mathcal{D}_{t,c} \\
                    0, & \text{otherwise}
                  \end{array}\right.
\end{align}

\subsection{wAR-RLS: The Closed-Form Solution}

Substituting (\ref{eq:l3}), (\ref{eq:fK}), (\ref{eq:DfKP}), and (\ref{eq:DfKQ2}) into (\ref{eq:f}), it follows that
\begin{align}
f=\argmin\limits_{f\in\mathcal{H}_K} &(\mathbf{y}^T-\boldsymbol{\alpha}^TK)E(\mathbf{y}-K\boldsymbol{\alpha})+ \sigma \boldsymbol{\alpha}^T K\boldsymbol{\alpha} \nonumber \\
& +  \boldsymbol{\alpha}^TK(\lambda_PM_0+\lambda_Q M)K\boldsymbol{\alpha} \label{eq:f3}
\end{align}
Setting the derivative of the objective function above to 0 leads to
\begin{align}
\boldsymbol{\alpha}=[(E+\lambda_P M_1+\lambda_Q M)K+\sigma I]^{-1}E\mathbf{y} \label{eq:alpha}
\end{align}

\subsection{wAR-SVM Solution}

Substituting (\ref{eq:l4}), (\ref{eq:fK}), (\ref{eq:DfKP}), and (\ref{eq:DfKQ2}) into (\ref{eq:f}), then $\boldsymbol{\alpha}$ in (\ref{eq:f2}) can be re-expressed as:
\begin{align}
\boldsymbol{\alpha}&=\argmin\limits_{\boldsymbol{\alpha}\in R^{n+m_l+m_u}\atop \boldsymbol{\xi}\in R^{n+m_l}}\sum_{i=1}^{n+m_l}E_{ii}\xi_i + \sigma \boldsymbol{\alpha}^T K\boldsymbol{\alpha} \nonumber \\
& \qquad\qquad \qquad \quad + \boldsymbol{\alpha}^TK(\lambda_PM_0+\lambda_Q M)K\boldsymbol{\alpha} \label{eq:fHinge}\\
\text{s.t.}&\quad y_i\left[\sum_{j=1}^{n+m_l+m_u}\alpha_j K(\mathbf{x}_i,\mathbf{x}_j)+b\right]\ge 1-\xi_i\nonumber\\
&\quad \xi_i\ge 0,\quad i=1,...,n+m_l \nonumber
\end{align}
Define
\begin{align*}
\boldsymbol{\beta}&=[\boldsymbol{\alpha};\quad  \boldsymbol{\xi};\quad  b]\\
\mathbf{f}&=[\mathbf{0}^{1\times (n+m_l+m_u)}\  w_{s,1}\ \cdots\ w_{s,n} \ w_tw_{t,1}\ \cdots\ w_tw_{t,m_l} \  0]\\
H&=\left[\begin{array}{cc}
          \sigma K+K(\lambda_PM_0+\lambda_QM)K &\mathbf{0}^{(n+m_l+m_u)\times(n+m_l+1)} \\
           \mathbf{0}^{(n+m_l+1)\times(n+m)} & \mathbf{0}^{(n+m_l+1)\times(n+m_l+1)}
         \end{array}\right]\\
A&=-[A'\quad \mathbf{I}^{(n+m_l)\times(n+m_l)} \quad \mathbf{y}]\\
B&=diag([\mathbf{0}^{1\times(n+m_l+m_u)}\quad \mathbf{1}^{1\times (n+m_l)}\quad 0])\\
\mathbf{b}&=-\mathbf{1}^{(n+m_l)\times 1}
\end{align*}
where $A'\in R^{(n+m_l)\times(n+m_l)}$ and $A'_{i,j}=\mathbf{y}_iK_{i,j}$, $\mathbf{0}^{1\times(n+m_l+m_u)}\in R^{1\times(n+m_l+m_u)}$ is a vector of all zeros, $\mathbf{1}^{1\times (n+m_l)}\in R^{1\times (n+m_l)}$ is a vector of all ones, and $\mathbf{I}^{(n+m_l)\times(n+m_l)}\in R^{(n+m_l)\times (n+m_l)}$ is the identity matrix.

Then, solving for $\boldsymbol{\alpha}$ and $b$ in (\ref{eq:fHinge}) is equivalent to solving for $\boldsymbol{\beta}$ below:
\begin{align}
\boldsymbol{\beta}&=\argmin\limits_{\boldsymbol{\beta}\in R^{2n+2m_l+m_u+1}} \boldsymbol{\beta}^TH\boldsymbol{\beta}+\mathbf{f}\boldsymbol{\beta} \label{eq:wAR-SVM} \\
\text{s.t.} & \quad A\cdot \boldsymbol{\beta}\le \mathbf{b} \nonumber\\
& \quad B\cdot \boldsymbol{\beta}\ge 0 \nonumber
\end{align}
which can be easily done using quadratic programming.

In summary, the pseudo code for wAR-RLS and wAR-SVM is shown in the first part of Algorithm~1.

\begin{algorithm}[htpb] %\DontPrintSemicolon
\KwIn{$n$ labeled source domain samples, $\{\mathbf{x}_i,y_i\}_{i=1}^n$;\\
\hspace*{10mm} $m_l$ labeled target domain samples, $\{\mathbf{x}_j,y_j\}_{j=n+1}^{n+m_l}$;\\
\hspace*{10mm} $m_u$ unlabeled target domain samples, $\{\mathbf{x}_j\}_{j=n+m_l+1}^{n+m_l+m_u}$;\\
\hspace*{10mm} Parameters $w_t$, $\sigma$, $\lambda_P$, and $\lambda_Q$;\\
\hspace*{10mm} $k$, number of unlabeled target domain samples to label.}
\KwOut{$\{y_j'\}_{j=n+m_l+1}^{n+m_l+m_u}$, estimated labels of the $m_u$ unlabeled target domain samples;\\
\hspace*{12mm} Indices of $k$ target domain samples to label.}
\tcp{wAR begins}
Compute $w_{s,i}$ and $w_{t,i}$ by (\ref{eq:ws}) and (\ref{eq:wt})\;
Compute the kernel matrix $K$\;
Construct $\{y_j\}_{j=n+m_l+1}^{n+m_l+m_u}$, pseudo labels for the $m_u$ unlabeled target domain samples, using the estimates from the previous iteration, or build another classifier (e.g., a basic SVM) to estimate the pseudo labels if this is the first iteration\;
Construct $\mathbf{y}$ in (\ref{eq:y}), $E$ in (\ref{eq:E}), $M_0$ in (\ref{eq:M0}), and $M$ in (\ref{eq:M})\;
Compute $\boldsymbol{\alpha}$ by (\ref{eq:alpha}) for wAR-RLS, or $\boldsymbol{\alpha}$ and $b$ by (\ref{eq:wAR-SVM}) for wAR-SVM\;
Compute $\{f(\mathbf{x}_j)\}_{j=n+m_l+1}^{n+m_l+m_u}$ by (\ref{eq:f2}) for wAR-RLS, or by (\ref{eq:f3}) for wAR-SVM\;
\textbf{Return} $\{y_j'\}_{j=n+m_l+1}^{n+m_l+m_u}$, where $y_j'=sign(f(\mathbf{x}_j))$\;
\tcp{wAR ends; AL begins}
Construct $J_d=\{j|y_j\neq y_j', n+m_l+1\le j\le n+m_l+m_u\}$\;
Sort $J_d$ in ascending order according to $|f(\mathbf{x}_j)|$, $j\in J_d$\;
Construct $J_s=\{j|y_j=y_j', n+m_l+1\le j\le n+m_l+m_u\}$\;
Sort $J_s$ in ascending order according to $|f(\mathbf{x}_j)|$, $j\in J_s$\;
Concatenate $J_d$ and $J_s$ to form an ordered set $J=\{J_d, J_s\}$\;
\textbf{Return} The first $k$ elements in $J$.\\
\tcp{AL ends}
\caption{The active weighted adaptation regularization (AwAR) algorithm.} \label{alg:AwAR}
\end{algorithm}

\section{Active Weighted Adaptation Regularization (AwAR)} \label{sect:AwAR}

As mentioned in the Introduction, wAR can be integrated with AL \cite{Settles2009} for better performance. AL tries to select the most informative samples to label so that a given learning performance can be achieved with less labeling effort. The key problem in using AL is estimating which of the data samples are the most informative. There are many different heuristics for this purpose \cite{Settles2009}. In this paper we select the most volatile and uncertain ones as the most informative ones. More sophisticated approaches will be studied in our future research\footnote{We attempted the active learning approaches in \cite{Chakraborty2015,Huang2014} but failed to observe better performance than the method proposed in this section.}.

\subsection{Active Learning} \label{sect:AL}

Our AL for identifying the $k$ most informative samples is a two-step procedure: the first step identifies the most volatile unlabeled target domain samples, and the second step further selects the $k$ most uncertain ones from them.

Recall that at the beginning of wAR we obtain $\{y_j\}_{j=n+m_l+1}^{n+m_l+m_u}$, the pseudo labels for unlabeled target domain samples, from the previous iteration, and finally we output $\{y_j'\}_{j=n+m_l+1}^{n+m_l+m_u}$, the updated estimates of these labels. If $y_j'$ is different from $y_j$ for a certain sample, then there is evidence that that sample is volatile, probably because it is close to the decision boundary. According to the volatility of the unlabeled target domain samples, we partition them into two groups: $J_d=\{j|y_j\neq y_j', n+m_l+1\le j\le n+m_l+m_u\}$ and $J_s=\{j|y_j=y_j', n+m_l+1\le j\le n+m_l+m_u\}$. Samples in $J_d$ are more volatile than those in $J_s$, and hence they are better candidates for labeling.

We further rank the uncertainties of the samples in $J_d$ by their closeness to the current decision boundary: a sample closer to the decision boundary means the classifier has more uncertainty about its class, and hence we should select it for labeling in the next iteration. To do this, we first sort $J_d$ in ascending order according to $|f(\mathbf{x}_j)|$. Since a smaller $|f(\mathbf{x}_j)|$ means a closer distance to the decision boundary and hence higher uncertainty, we select the first $k$ samples in $J_d$ for labeling in the next iteration. If $k$ is larger than the number of samples in $J_d$, then we also sort $J_s$ in ascending order according to $|f(\mathbf{x}_j)|$ and select the first $k-|J_d|$ samples from it.

\subsection{The Complete AwAR Algorithm}

The complete AwAR algorithm is given in Algorithm~\ref{alg:AwAR}. We denote the one based on wAR-RLS as AwAR-RLS, and the one based on wAR-SVM as AwAR-SVM. In each algorithm, we first use wAR to classify the unlabeled target domain samples, and then AL to identify $k$ such samples that are most volatile and uncertain. AwAR-RLS and AwAR-SVM can easily be embedded into an iterative procedure (Section~\ref{sect:process}) so that $k$ target domain samples are labeled in each iteration until the maximum number of iterations is reached, or the desired classification performance is achieved.

\subsection{Make Use of the Extra Channels}

In Algorithm~1, we assume the source and target domains have consistent features, i.e., the old and new headsets have same channels so that the features extracted from them have the same dimensionality and meaning. This also works if the old headset has more channels, but it includes all channels in the new headset, in which case only the common channels are used in feature extraction. However, things become more complicated if the new headset has channels that are not included in the old headset. We can again use the common channels for feature extraction and then apply Algorithm~1, but there is information loss if the extra channels in the new headset are completely ignored. We next propose a solution for this problem.

The extra channels are difficult to use in wAR, because the target domain does not contain them. However, it is possible to use them in AL, as shown in Algorithm~2, which can be used to replace the AL part in Algorithm~1. Algorithm~2 still consists of two steps. The first step identifies the most volatile unlabeled target domain samples, which is the same as that in the original AL algorithm. The second step ranks the uncertainties of the unlabeled samples by incorporating the uncertainty information from all channels (common channels plus extra channels). For that we first build a separate classifier using features extracted from all channels and trained from only the $m_l$ labeled samples. For each unlabeled sample, we compute the sum of two signed distances: 1) the distance from the decision boundary determined by this additional classifier, and 2) the distance from the decision boundary determined by wAR. The smaller the sum, the larger the uncertainty. We then return the top $k$ unlabeled samples that are volatile and most uncertain.

\begin{algorithm}[!h] %\DontPrintSemicolon
\tcp{wAR ends; AL begins}
Design another classifier, e.g., a SVM, to classify the $m_u$ unlabeled target domain samples using features from all channels; denote the signed distances to its decision boundary as $\{g(\mathbf{x}_j)\}_{j=n+m_l+1}^{n+m_l+m_u}$\;
$J_d=\{j|y_j\neq y_j', n+m_l+1\le j\le n+m_l+m_u\}$\;
Sort $J_d$ in ascending order according to $|f(\mathbf{x}_j)+g(\mathbf{x}_j)|$, $j\in J_d$\;
$J_s=\{j|y_j=y_j', n+m_l+1\le j\le n+m_l+m_u\}$\;
Sort $J_s$ in ascending order according to $|f(\mathbf{x}_j)+g(\mathbf{x}_j)|$, $j\in J_s$\;
Concatenate $J_d$ and $J_s$ to form the ordered set $J=\{J_d, J_s\}$\;
\textbf{Return} The first $k$ elements in $J$.\\
\tcp{AL ends}
\caption{The active learning (AL) algorithm for making use of extra channels in the target domain.} \label{alg:AL}
\end{algorithm}

\section{Experiments and Discussions} \label{sect:experiments}

Experimental results are presented in this section to compare wAR-RLS, wAR-SVM, AwAR-RLS, and AwAR-SVM with several other algorithms.

\subsection{Experiment Setup}

\begin{figure}[htpb]\centering
\subfigure[]{\label{fig:T}     \includegraphics[width=.23\linewidth]{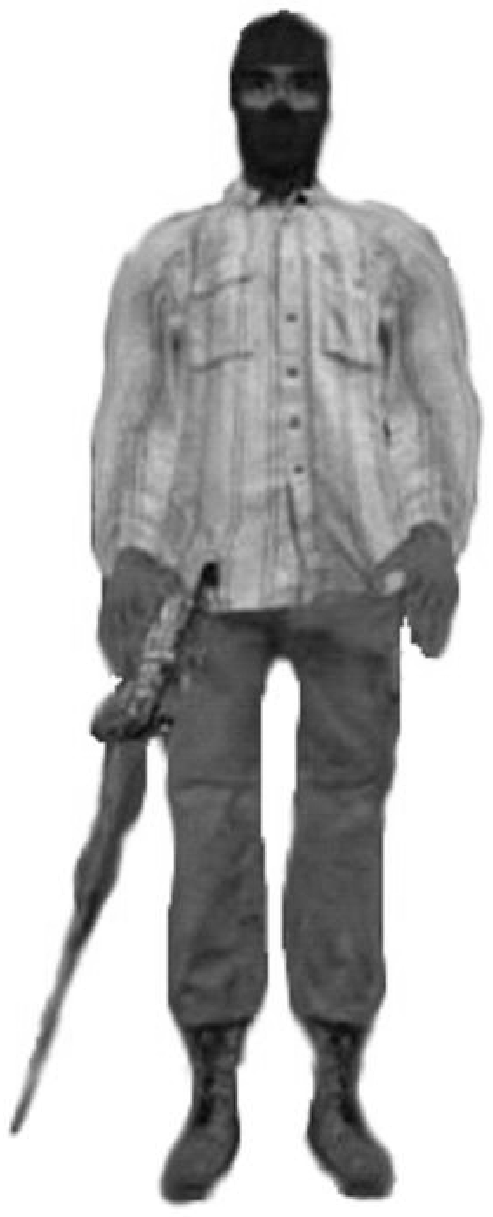}}
\subfigure[]{\label{fig:NT}     \includegraphics[width=.20\linewidth]{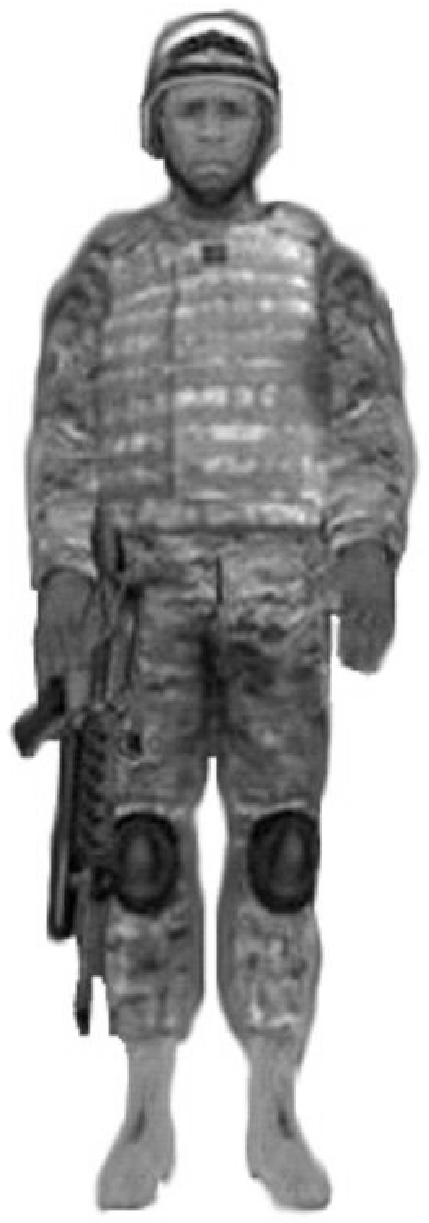}}
\caption{Example images of (a) a target; (b) a non-target.} \label{fig:TNT}
\end{figure}

We used data from a VEP oddball task \cite{Ries2014}. In this task, image stimuli were presented to subjects at a rate of 0.5 Hz (one image every two seconds). The images presented were either an enemy combatant [target; an example is shown in Fig.~\ref{fig:T}] or a U.S. Soldier [non-target; an example is shown in Fig.~\ref{fig:NT}]. The subjects were instructed to identify each image as being target or non-target with a unique button press as quickly, but as accurately, as possible. There were a total of 270 images presented to each subject, of which 34 were targets. The experiments were approved by the U.S. Army Research Laboratory (ARL) Institutional Review Board (Protocol \# 20098-10027). The voluntary, fully informed consent of the persons used in this research was obtained as required by federal and Army regulations \cite{USArmy,USDoD}. The investigator adhered to Army policies for the protection of human subjects.

Eighteen subjects participated in the experiments, which lasted on average 15 minutes. Data from four subjects were not used due to data corruption or poor responses. Signals were recorded with  three  different EEG  headsets, including a wired 64-channel ActiveTwo\footnote{http://www.biosemi.com/products.htm} system (sample rate set to 512Hz) from BioSemi, a wireless 9-channel 256Hz B-Alert X10 EEG Headset System\footnote{http://www.advancedbrainmonitoring.com/xseries/x10/} from Advanced Brain Monitoring (ABM), and a wireless 14-channel 128Hz EPOC headset\footnote{https://emotiv.com/epoc.php} from Emotiv. We considered  switching  between BioSemi and Emotiv headsets, and between BioSemi and ABM headsets, respectively. Switching between Emotiv and ABM headsets was not considered because they have too few common channels.

\subsection{Preprocessing and Feature Extraction}

We used EEGLAB \cite{Delorme2014} for EEG signal preprocessing and feature extraction. Raw amplitude features were used in this study. The performances of AwAR-RLS and AwAR-SVM on other feature sets are studied later in this section.

For switching between BioSemi and Emotiv headsets, we used their 14 common channels (AF3, AF4, F3, F4, F7, F8, FC5, FC6, O1, O2, P7, P8, T7, T8). For switching between BioSemi and ABM headsets, we used their nine common channels (C3, C4, Cz, F3, F4, Fz, P3, P4, POz). For each headset, we first band-passed the EEG signals to [1, 50] Hz, then downsampled them to 64 Hz, performed average reference, and next epoched them to the $[0, 0.7]$ second interval timelocked to stimulus onset. We removed mean baseline from each channel in each epoch and removed epochs with incorrect button press responses\footnote{Button press responses were not recorded for the ABM headset, so we used all epochs from it.}. The final numbers of epochs from the 14 subjects are shown in Table~\ref{tab:epoch}. Observe that there is significant class imbalance for all headsets; that's why we need to use $w_{s,i}$ and $w_{t,i}$ in (\ref{eq:f}) to balance the two classes in both domains.

\begin{table*}[htpb] \centering \setlength{\tabcolsep}{1mm}
\caption{Number of epochs for each subject after preprocessing. The numbers of target epochs are given in the parentheses.}   \label{tab:epoch}
\begin{tabular}{l|cccccccccccccc}   \hline
   Subject  &  1&2&3&4&5&6&7&8&9&10&11&12&13 &14 \\ \hline
   BioSemi &  241(26)&260(24)& 257(24) & 261(29)& 259(29)& 264(30)& 261(29) & 252(22)& 261(26)& 259(29)& 267(32)& 259(24)&261(25)& 269(33)\\
  Emotiv  &263(28) &  265(30) &  266(30)& 255(23)& 264(30)& 263(32)& 266(30)&252(22)& 261(26)& 266(29)& 266(32)& 264(33) & 261(26)& 267(31)\\
  ABM & 270(34) & 270(34) & 235(30) & 270(34) & 270(34)&270(34)&270(34)&270(33)&270(34)&239(30)&270(34)&270(34)&251(31)&270(34)\\
  \hline
\end{tabular}
\end{table*}

Each [0, 0.7] second epoch contains 45 raw EEG magnitude samples. The concatenated feature vector has hundreds of dimensions. To reduce the dimensionality, we combined concatenated feature vectors from the old and new headsets, performed a simple principal component analysis (PCA), and took only the scores for the first 20 principal components (PCs). We then normalized each feature dimension separately to $[0, 1]$ for each subject.

\subsection{Evaluation Process and Performance Measures} \label{sect:process}

\begin{figure}[htpb]\centering
 \includegraphics[width=\linewidth]{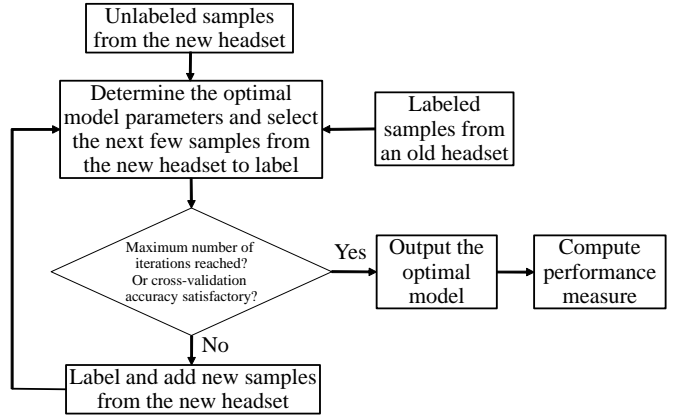} \caption{Flowchart of the evaluation process.} \label{fig:flowchart}
\end{figure}

Although we know the labels of all EEG epochs from all headsets for each subject, we simulate a different scenario, as shown in Fig.~\ref{fig:flowchart}: all EEG epochs from the old headset are labeled, but none of the epochs from the new headset is initially labeled. Our approach is to iteratively label some epochs from the new headset, and then to build a classifier to label the rest of the epochs. The goal is to achieve the highest classification accuracy for the epochs from the new headset, with as few labeled epochs as possible.

The following three performance measures were used:
\begin{enumerate}
\item False positive rate (FPR), which is the number of false positives (the number of non-targets which were mistakenly classified as targets) divided by the number of true negatives (non-targets).
\item False negative rate (FNR), which is the number of false negatives (the number of targets which were mistakenly classified as non-targets) divided by the number of true positives (targets).
\item Balanced classification accuracy (BCA), which is the average of classification accuracies on the positive (target) class and the negative (non-target) class. It can be shown that $BCA=1-(FPR+FNR)/2$.
\end{enumerate}

\subsection{Algorithms}

We compared the performances of wAR-RLS, wAR-SVM, AwAR-RLS and AwAR-SVM with three other algorithms:
\begin{enumerate}
\item Baseline (BL), which is a simple iterative procedure: in each iteration we randomly select a few unlabeled training samples collected using the new headset, ask the subject to label them, add them to the labeled training dataset, and then train an SVM classifier by 5-fold cross-validation. We iterate until the maximum number of iterations is reached.
\item The simple TL (TL) algorithm introduced in \cite{drwuSMC2014}, which is very similar to BL, except that in each iteration it combines labeled samples from the old and new headsets in building an SVM classifier and then applies it to the unlabeled samples from the new headset.
\item  The active TL (ATL) algorithm introduced in \cite{drwuSMC2014}, which adds AL to the above TL: instead of randomly selecting unlabeled samples from the new headset to label, it selects those closest to the SVM decision boundary.
\end{enumerate}

Weighted LIBSVM \cite{LIBSVM} with a linear kernel was used as the classifier in BL, TL, ATL, wAR-SVM, and AwAR-SVM. Grid search was used to determine the optimal penalty parameter in LIBSVM for BL, TL and ATL. We chose $w_t=2$ in wAR-RLS, wAR-SVM, AwAR-RLS and AwAR-SVM to give the labeled target domain samples more weights, and $\sigma=0.1$ and $\lambda_P=\lambda_Q=10$, following the practice in \cite{Long2014}. In Section~\ref{sect:robustness} we present robustness analysis for AwAR-RLS and AwAR-SVM to $\sigma$, $\lambda_P$ and $\lambda_Q$, and show that AwAR-RLS and AwAR-SVM are insensitive to them. Because there are labeled target domain samples, cross-validation could also be used to optimize these parameters. This will be considered in our future research.

\subsection{Experimental Results} \label{sect:results}

All seven algorithms started with zero labeled samples from the new headset. In each iteration, five new EEG epochs were labeled and added to the training dataset. For BL, TL, wAR-RLS and wAR-SVM, these five were the same and were selected randomly from unlabeled samples. For ATL, AwAR-RLS and AwAR-SVM, these five were selected by their respective AL algorithms, so generally they were different in different algorithms.

\begin{figure}[!h]\centering
\subfigure[]{\label{fig:B2Aavg}\includegraphics[width=\linewidth]{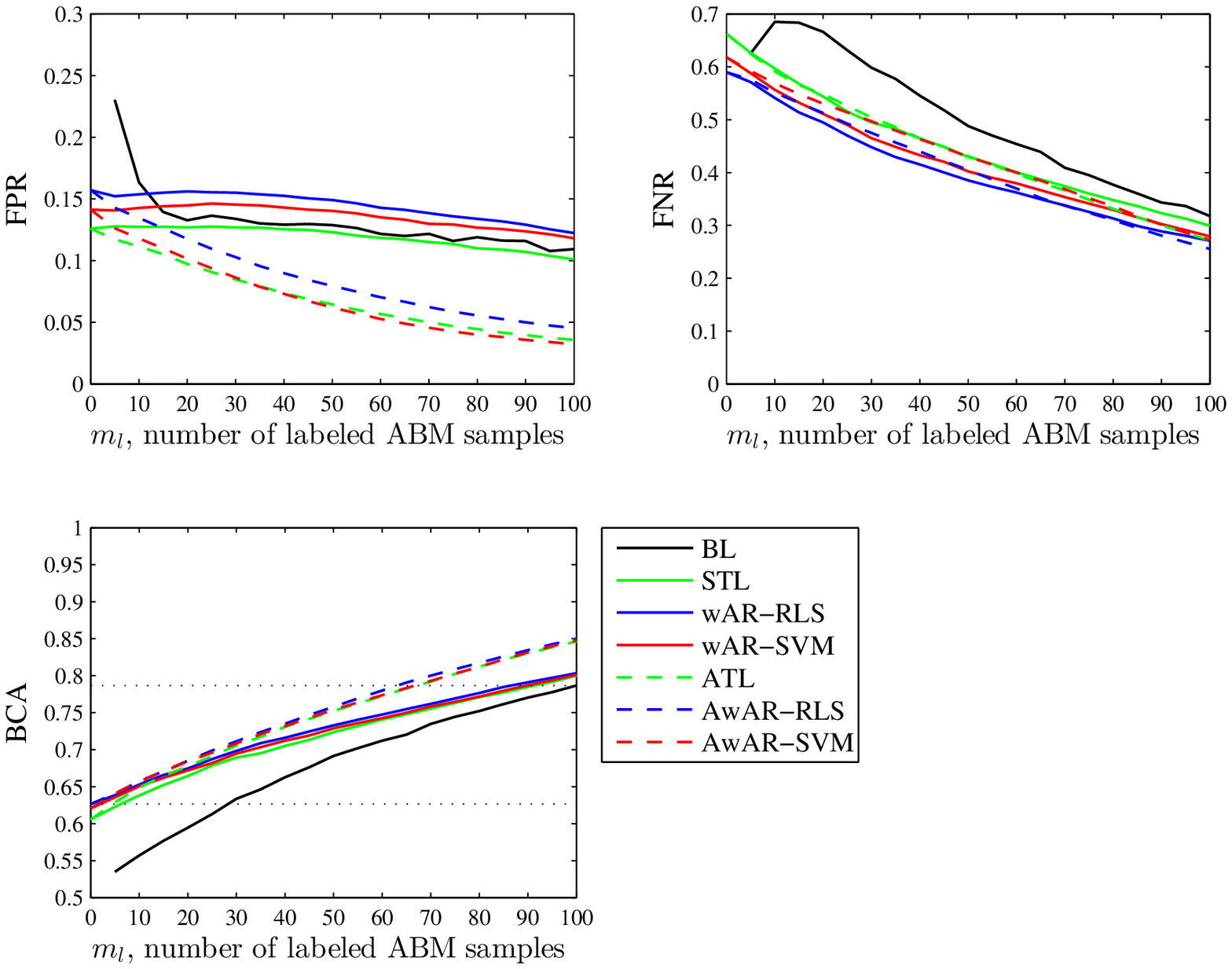}}\hfill
\subfigure[]{\label{fig:A2Bavg}\includegraphics[width=\linewidth]{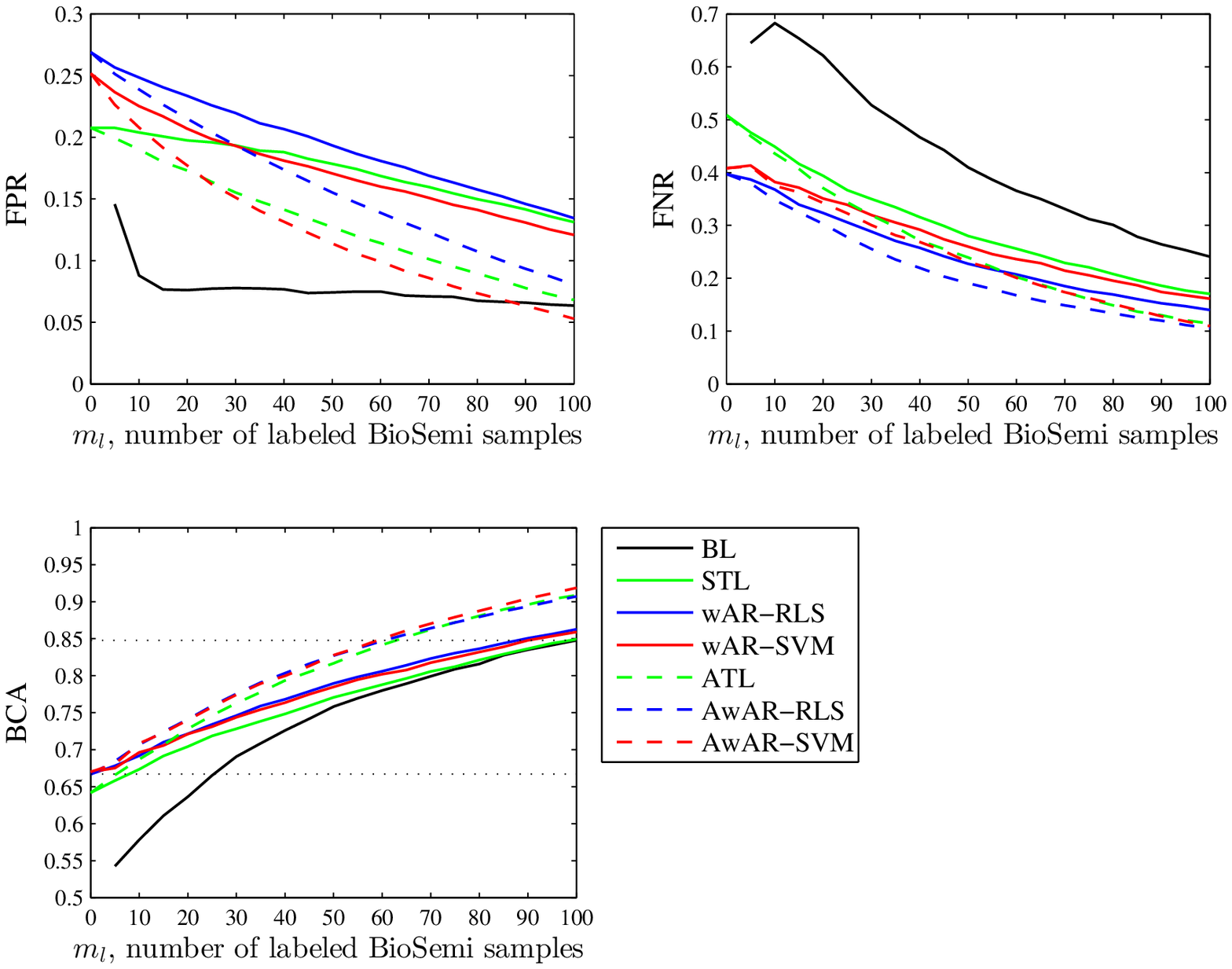}}
\caption{Average performances of the seven algorithms across the 14 subjects across the BioSemi and ABM headsets. (a) Switching from BioSemi headset to ABM headset; (b) switching from ABM headset to BioSemi headset.} \label{fig:avgP2}
\end{figure}

To cope with randomness in these methods, each of them was repeated 30 times and the average results are shown. Because the AL-based algorithms are deterministic, we introduced randomness by randomly selecting (without replacement) 200 epochs from the old headset as data in the source domain, before running the seven algorithms. The average performances of the seven algorithms across the 14 subjects for the four switching scenarios are shown in Figs.~\ref{fig:avgP} and \ref{fig:avgP2}. Observe that:

\begin{figure}[!h]\centering
\subfigure[]{\label{fig:B2Eavg}\includegraphics[width=\linewidth]{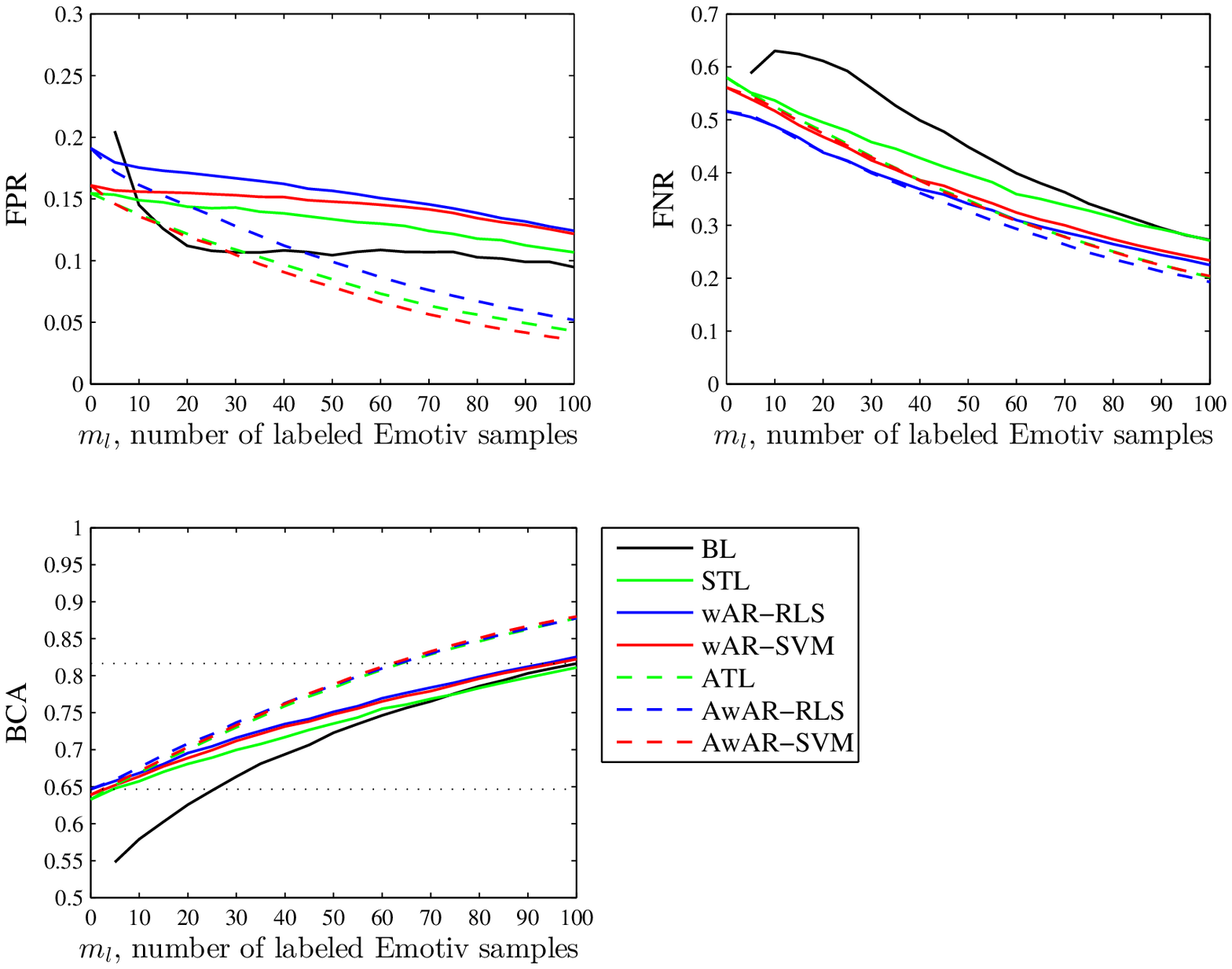}}\hfill
\subfigure[]{\label{fig:E2Bavg}\includegraphics[width=\linewidth]{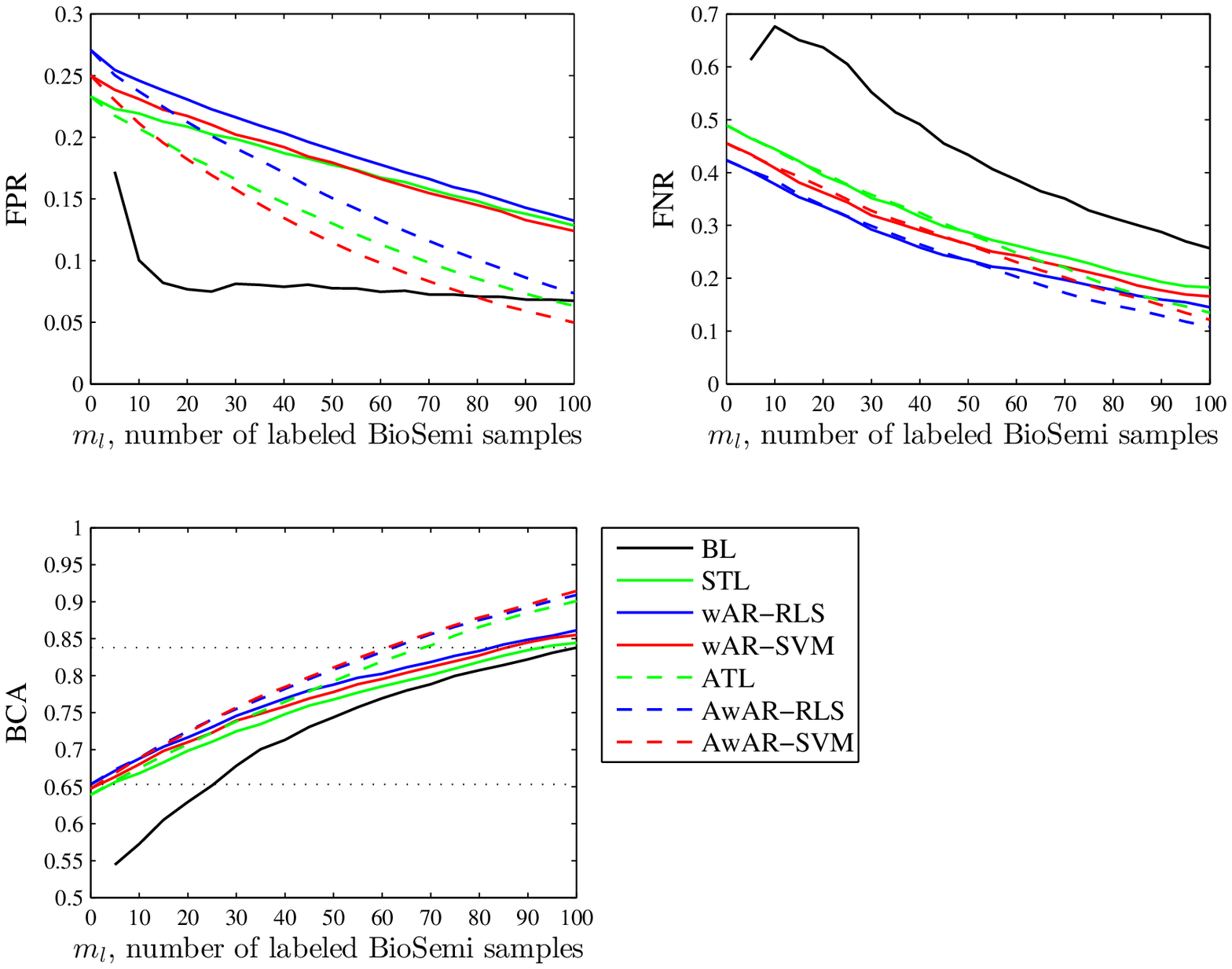}}
\caption{Average performances of the seven algorithms across the 14 subjects across the BioSemi and Emotiv headsets. (a) Switching from BioSemi headset to Emotiv headset; (b) switching from Emotiv headset to BioSemi headset.} \label{fig:avgP}
\end{figure}

\begin{enumerate}
\item Generally, the performance of BL increases as more samples from the new headset are labeled and added; however, it cannot build a model when there are no labeled samples at all from the new headset (observe that the first point on the BL curve is missing in every subfigure). On the contrary, without any labeled samples from the new headset, all other TL or wAR-based methods can build a model which has over 50\%, many times much higher, BCA for most subjects, because they can transfer useful knowledge from the old headset to the new one. More specifically, the first point on the TL (or ATL) curve in each subfigure represents the BCA when the best classifier learned from the old headset is applied directly to the new headset. Observe that it is better than 50\% (random guess) for most subjects. However, better BCAs can be obtained with wAR and AwAR.

\item Generally, all six TL or wAR-based methods outperform BL, which is expected, as TL and wAR get additional data from the old headset.

\item AwAR-RLS almost always achieves better performance (in terms of FPR, FNR, and BCA) than wAR-RLS, and AwAR-SVM almost always achieves better performance than wAR-SVM. The average performance improvements of AwAR-RLS over wAR-RLS, and AwAR-SVM over wAR-SVM, are evident for all four scenarios, as shown in Figs.~\ref{fig:avgP} and \ref{fig:avgP2}. This verifies our conjecture that integrating AL with wAR can further improve the performance of wAR.

\item As shown in Figs.~\ref{fig:avgP} and \ref{fig:avgP2}, among the three AL methods (ATL, AwAR-RLS and AwAR-SVM), AwAR-SVM almost always have the smallest FPR, and AwAR-RLS almost always have the smallest FNR. AwAR-RLS and AwAR-SVM have higher BCAs than ATL when $m_l$ is small, but they become closer as $m_l$ increases. AwAR-RLS and AwAR-SVM have better performance than ATL, because they use more sophisticated wAR algorithms. As an evidence, Figs.~\ref{fig:avgP} and \ref{fig:avgP2} also show that wAR-RLS and wAR-SVM achieve better performance than TL.

\item Generally, wAR-RLS has similar performance to wAR-SVM, and AwAR-RLS also has similar performance to AwAR-SVM. However, since wAR-RLS and AwAR-RLS can be trained several times faster than wAR-SVM and AwAR-SVM, they are the preferred methods to use. This is also consistent with the observations in \cite{Long2014}.
\end{enumerate}

\subsection{Statistical Analysis}

We also performed comprehensive statistical tests to check if the BCA differences among the algorithms were statistically significant. To assess overall performance differences among all the algorithms, a measure called the area-under-performance-curve (AUPC) \cite{drwuRSVP2016} was calculated. The AUPC is the area under the curve of the BCA values plotted at each of the 30 random runs and is normalized to $[0, 1]$. Larger AUPC values indicate better overall classification performance.

First, we used Friedman's test, a two-way non-parametric Analysis of Variance (ANOVA) where column effects are tested for significant differences after adjusting for possible row effects. We treated the algorithm type (BL, TL, wAR-RLS, wAR-SVM, ATL, AwAR-RLS, AwAR-SVM) as the column effects, with subjects as the row effects. Each combination of algorithm and subject had 30 values corresponding to 30 random runs performed. Friedman's test showed statistically significant differences among the seven algorithms ($p=.0000$) across all four modes of transfer (BioSemi $\leftrightarrow$ ABM, Emotiv $\leftrightarrow$ BioSemi).

Then, non-parametric multiple comparison tests using Dunn's procedure \cite{Dunn1961,Dunn1964} were used to determine if the difference between any pair of algorithms was statistically significant, with a $p$-value correction using the False Discovery Rate method by \cite{Benjamini1995}. This test was performed for each mode of transfer, and the results are shown in Tables~\ref{tab:1}-\ref{tab:4}. Observe that in all cases, AL based methods (ATL, AwAR-SVM, AwAR-RLS) performed significantly better than the corresponding non-AL based methods. AwAR-RLS and AwAR-SVM always performed significantly better than BL, TL, wAR-RLS and wAR-SVM. Although AwAR-RLS and AwAR-SVM did not perform significantly better than ATL, the $p$-values were close to the threshold when switching from Emotiv to BioSemi (Table~\ref{tab:1}), and from ABM to BioSemi (Table~\ref{tab:4}). The BCA difference between AwAR-RLS and AwAR-SVM was always not statistically significant.

\begin{table}[ht] \centering \setlength{\tabcolsep}{1mm}
\caption{$p$-values of non-parametric multiple comparison of BCAs of the algorithms when switching from Emotiv to BioSemi.}   \label{tab:1}
\begin{tabular}{l|ccccccc}   \hline
     &  BL & STL & wAR-RLS & wAR-SVM & ATL & AwAR-RLS \\ \hline
   TL &  \textbf{.0000} &&&&&\\
  wAR-RLS  &\textbf{.0000}&\textbf{.0055}&&&&\\
  wAR-SVM & \textbf{.0000}&.1042&.1091&&&\\
  ATL&\textbf{.0000}&\textbf{.0000}&\textbf{.0000}&\textbf{.0000}&&\\
  AwAR-RLS&\textbf{.0000}&\textbf{.0000}&\textbf{.0000}&\textbf{.0000}&.0788&\\
  AwAR-SVM&\textbf{.0000}&\textbf{.0000}&\textbf{.0000}&\textbf{.0000}&.1297&.3572\\   \hline
\end{tabular}
\end{table}

\begin{table}[ht] \centering \setlength{\tabcolsep}{1mm}
\caption{$p$-values of non-parametric multiple comparison of BCAs of the algorithms when switching from BioSemi to Emotiv.}   \label{tab:2}
\begin{tabular}{l|ccccccc}   \hline
     &  BL & STL & wAR-RLS & wAR-SVM & ATL & AwAR-RLS \\ \hline
   TL &  \textbf{.0000} &&&&&\\
  wAR-RLS  &\textbf{.0000}&\textbf{.0299}&&&&\\
  wAR-SVM & \textbf{.0000}&.0882&.3117&&&\\
  ATL&\textbf{.0000}&\textbf{.0000}&\textbf{.0000}&\textbf{.0000}&&\\
  AwAR-RLS&\textbf{.0000}&\textbf{.0000}&\textbf{.0000}&\textbf{.0000}&.2731&\\
  AwAR-SVM&\textbf{.0000}&\textbf{.0000}&\textbf{.0000}&\textbf{.0000}&.2680&.4892\\   \hline
\end{tabular}
\end{table}

\begin{table}[ht] \centering \setlength{\tabcolsep}{1mm}
\caption{$p$-values of non-parametric multiple comparison of BCAs of the algorithms when switching from BioSemi to ABM.}   \label{tab:3}
\begin{tabular}{l|ccccccc}   \hline
     &  BL & STL & wAR-RLS & wAR-SVM & ATL & AwAR-RLS \\ \hline
   TL &  \textbf{.0000} &&&&&\\
  wAR-RLS  &\textbf{.0000}&.1478&&&&\\
  wAR-SVM & \textbf{.0000}&.2511&.3525&&&\\
  ATL&\textbf{.0000}&\textbf{.0019}&.0397&\textbf{.0160}&&\\
  AwAR-RLS&\textbf{.0000}&\textbf{.0001}&\textbf{.0038}&\textbf{.0011}&.2044&\\
  AwAR-SVM&\textbf{.0000}&\textbf{.0008}&\textbf{.0200}&\textbf{.0072}&.3781&.2808\\   \hline
\end{tabular}
\end{table}

\begin{table}[!ht] \centering \setlength{\tabcolsep}{1mm}
\caption{$p$-values of non-parametric multiple comparison of BCAs of the algorithms when switching from ABM to BioSemi.}   \label{tab:4}
\begin{tabular}{l|ccccccc}   \hline
     &  BL & STL & wAR-RLS & wAR-SVM & ATL & AwAR-RLS \\ \hline
   TL &  \textbf{.0000} &&&&&\\
  wAR-RLS  &\textbf{.0000}&\textbf{.0011}&&&&\\
  wAR-SVM & \textbf{.0000}&\textbf{.0171}&.1854&&&\\
  ATL&\textbf{.0000}&\textbf{.0000}&\textbf{.0002}&\textbf{.0000}&&\\
  AwAR-RLS&\textbf{.0000}&\textbf{.0000}&\textbf{.0000}&\textbf{.0000}&.0874&\\
  AwAR-SVM&\textbf{.0000}&\textbf{.0000}&\textbf{.0000}&\textbf{.0000}&.0504&.3808\\   \hline
\end{tabular}
\end{table}

In summary, we have demonstrated that AwAR-RLS and AwAR-SVM can significantly improve the BCA, given the same number of labeled samples from the new headset. In other words, given a desired BCA, these algorithms can significantly reduce the number of labeled samples from the new headset. For example, Figs.~\ref{fig:avgP} and \ref{fig:avgP2} show that on average, AwAR-RLS and AwAR-SVM can achieve the same BCA as BL, trained from 100 labeled samples from the new headset, using only 60 to 65 labeled samples. Figs.~\ref{fig:avgP} and \ref{fig:avgP2} also show that, without using any labeled samples from the new headset, on average AwAR-RLS and AwAR-SVM can achieve the same BCA as BL which is trained from about 25 labeled samples from the new headset.

\subsection{Make Use of the Extra Channels (ECs)}

In the above experiments, we have only used the common channels between the old and new headsets. This is fine if all channels of the new headset are included in the old headset; however, there is information loss if the new headset has channels that do not present in the old headset. For example, when switching from Emotiv to BioSemi, the extra $64-14=50$ channels are completely ignored, whereas they may contain valuable information.

In this subsection, we replace the AL part in Algorithm~1 by Algorithm~2 to make use of the extra channels, and the corresponding algorithms are denoted as AwAR-RLS-EC and AwAR-SVM-EC. Because this modification only affects AwAR-RLS and AwAR-SVM, we do not present results from STL, wAR-RLS and wAR-SVM since they are the same as those in the last subsection. However, for comparison purpose, we include BL and ATL. We also added another baseline algorithm (BL-EC), which is similar to BL in the last subsection but uses features extracted from all 64 BioSemi channels.

The average results across the 14 subjects are shown in Fig.~\ref{fig:avgP_EC}, and the results for the individual subjects are shown in the Appendix. Observe from Fig.~\ref{fig:avgP_EC} that by making use of the extra channels, BL-EC had better FPR, FNR and BCA than BL, AwAR-RLS-EC had better FPR, FNR and BCA than AwAR-RLS, and AwAR-SVM-EC also had better FPR, FNR and BCA than AwAR-SVM. In summary, Algorithm~2 indeed allowed us to exploit new information in the extra channels to improve performance.

\begin{figure}[htpb]\centering
\subfigure[]{\label{fig:E2Bavg_EC}\includegraphics[width=\linewidth]{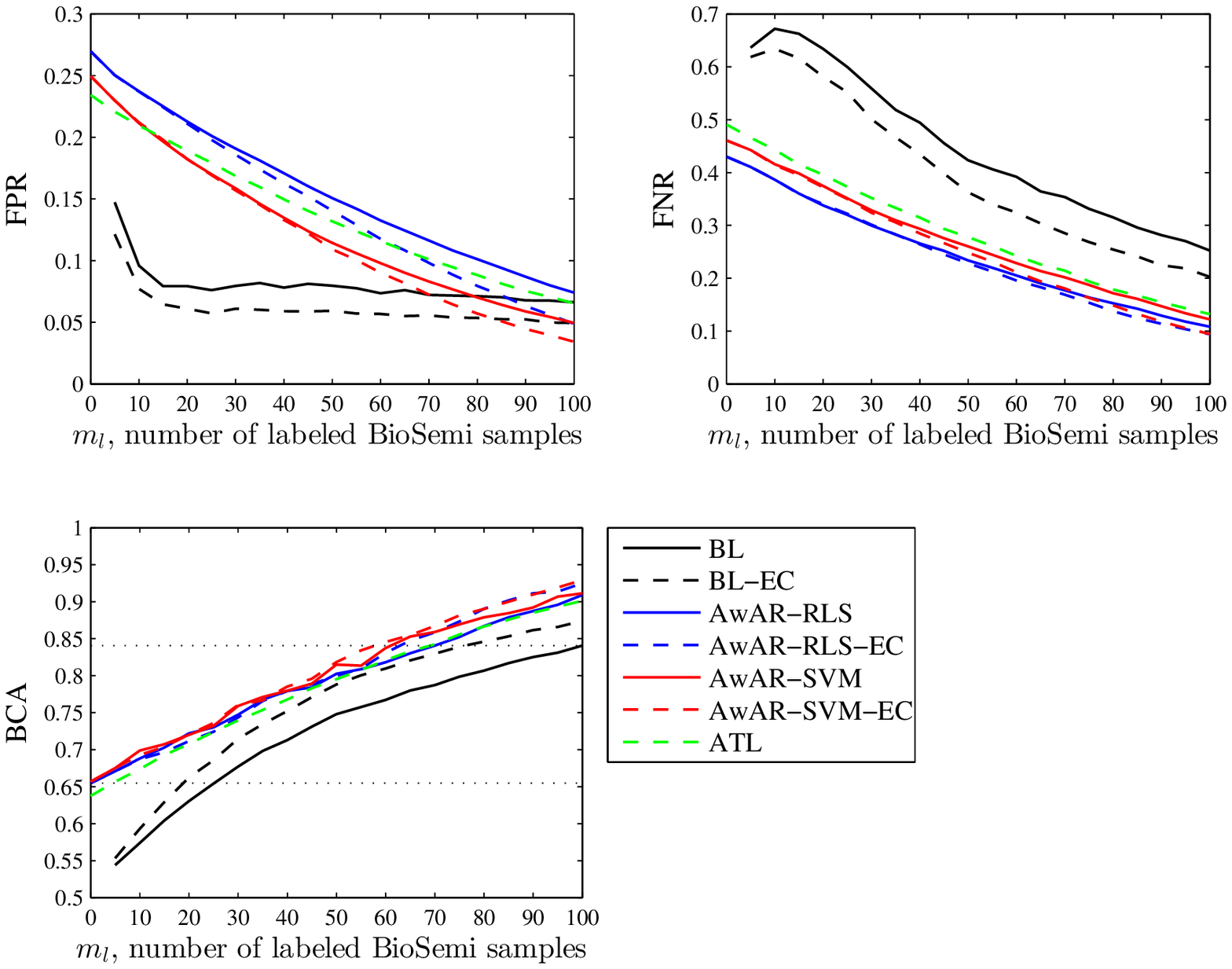}}
\subfigure[]{\label{fig:A2Bavg_EC}\includegraphics[width=\linewidth]{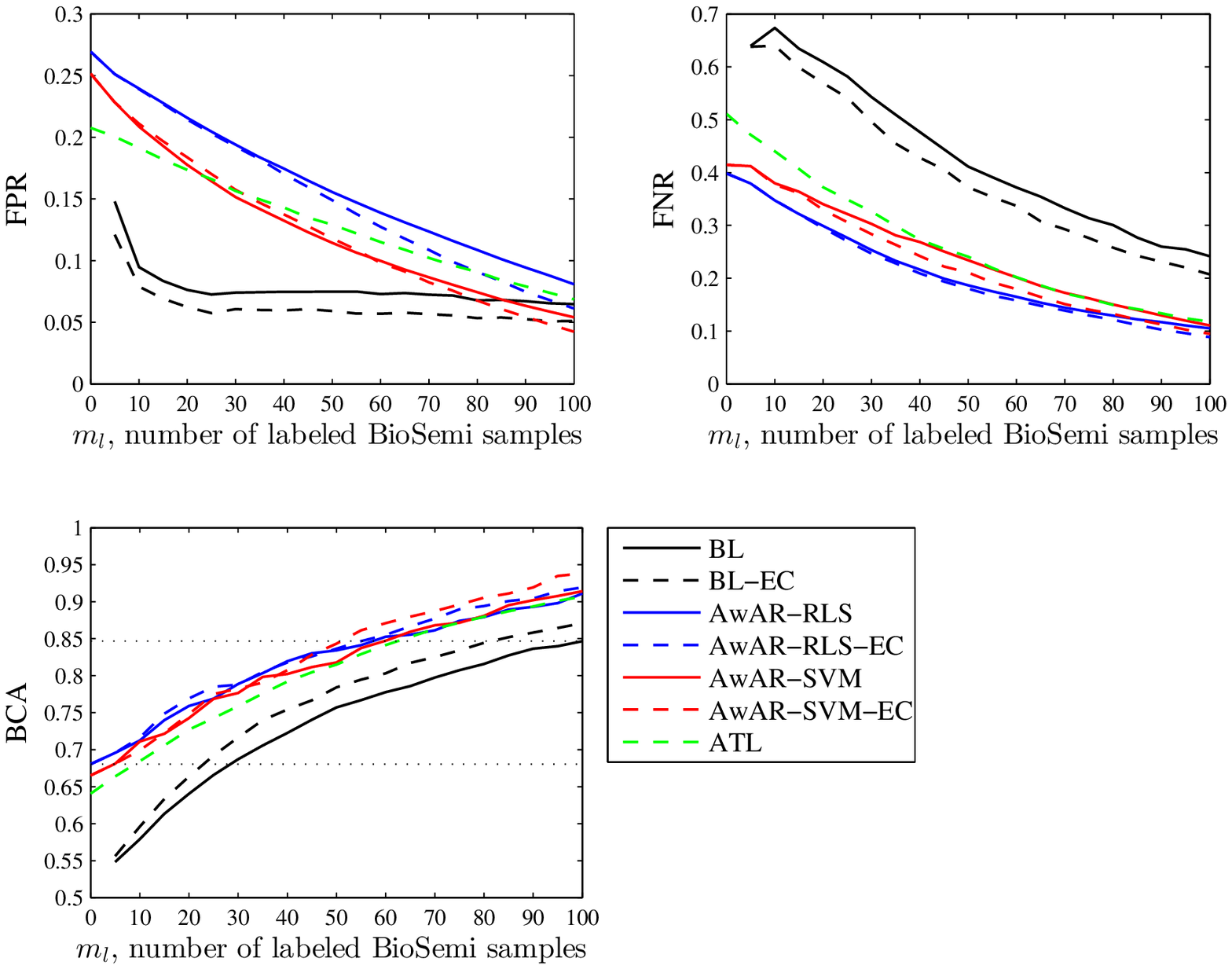}}
\caption{Average performances of the seven algorithms across the 14 subjects. (a) Switching from Emotiv headset to BioSemi headset; (b) switching from ABM headset to BioSemi headset.} \label{fig:avgP_EC}
\end{figure}

We also performed statistical tests to check if the BCA improvement with the extra channels were statistically significant. Friedman's test showed statistically significant difference among the six learning algorithms ($p=.0000$) across both modes of transfer (Emotiv $\rightarrow$ Biosemi, ABM $\rightarrow$ Biosemi). Dunn's procedure (Tables~\ref{tab:5}-\ref{tab:6}) showed that BL-EC was always statistically better than BL. AwAR-SVM-EC was statistically better than AwAR-SVM when switching from ABM to BioSemi. With the help of the extra channels, AwAR-SVM-EC had statistically better BCA than ATL when switching from Emotiv to BioSemi, and both AwAR-SVM-EC and AwAR-RLS-EC had statistically better BCAs than ATL when switching from ABM to BioSemi.

\begin{table}[!ht] \centering \setlength{\tabcolsep}{1mm}
\caption{$p$-values of non-parametric multiple comparison of the six algorithms when switching from Emotiv to BioSemi, with extra channels.}   \label{tab:5}
\begin{tabular}{l|ccccccc}   \hline
     &    &       &  AwAR& AwAR- &   AwAR& AwAR-\\
    &  BL & BL-EC &-RLS    & RLS-EC &-SVM    & SVM-EC \\\hline
   BL-EC &  \textbf{.0001} & &&&&\\
  AwAR-RLS  &\textbf{.0000}&\textbf{.0000}&&&&\\
  AwAR-RLS-EC & \textbf{.0000}&\textbf{.0000}&.1677&&&\\
  AwAR-SVM&\textbf{.0000}&\textbf{.0000}&.1531&.4636&&\\
  AwAR-SVM-EC&\textbf{.0000}&\textbf{.0000}&\textbf{.0167}&.1490&.1562&\\
  ATL &\textbf{.0000} & \textbf{.0000} & .4616 & .1467 & .1422 &\textbf{.0119}\\ \hline
\end{tabular}
\end{table}

\begin{table}[!ht] \centering \setlength{\tabcolsep}{1mm}
\caption{$p$-values of non-parametric multiple comparison of the seven algorithms when switching from ABM to BioSemi, with extra channels.}   \label{tab:6}
\begin{tabular}{l|ccccccc}   \hline
     &     &       & AwAR& AwAR-  &  AwAR& AwAR-\\
     &  BL & BL-EC &-RLS   & RLS-EC &-SVM    & SVM-EC\\ \hline
   BL-EC &  \textbf{.0000} &&&&&\\
  AwAR-RLS  &\textbf{.0000}&\textbf{.0000}&&&&\\
  AwAR-RLS-EC & \textbf{.0000}&\textbf{.0000}&.1669&&&\\
  AwAR-SVM&\textbf{.0000}&\textbf{.0000}&.0443&\textbf{.0032}&&\\
  AwAR-SVM-EC&\textbf{.0000}&\textbf{.0000}&.1950&.4450&\textbf{.0046}&\\
  ATL & \textbf{.0000}&\textbf{.0000}&\textbf{.0008}&\textbf{.0000}&.0839&\textbf{.0000}\\\hline
\end{tabular}
\end{table}

\subsection{Robustness Analysis} \label{sect:robustness}

In this subsection we study the robustness of AwAR-RLS and AwAR-SVM to three different factors: the number of linear PC features, the feature sets extracted using different methods, and the parameters $\sigma$ and $\lambda_P$ ($\lambda_Q$). To save space, we only show the BCA results when switching from BioSemi to ABM. Similar results were obtained from other switching scenarios.

\begin{figure}[htpb]\centering
\includegraphics[width=\linewidth]{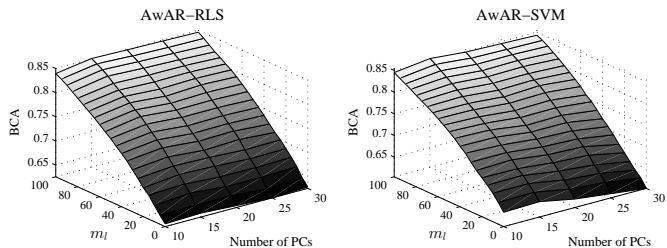}
\caption{Average BCAs of AwAR-RLS and AwAR-SVM for different number of linear PCs, when switching from BioSemi to ABM.} \label{fig:numPCs}
\end{figure}

The average BCAs of AwAR-RLS and AwAR-SVM for different number of linear PCs are shown in Fig.~\ref{fig:numPCs}. Observe that AwAR-RLS and AwAR-SVM are very robust to the number of PCs. 20 PCs were used in this paper mainly for the computational cost consideration.

\begin{figure}[htpb]\centering
\includegraphics[width=.9\linewidth]{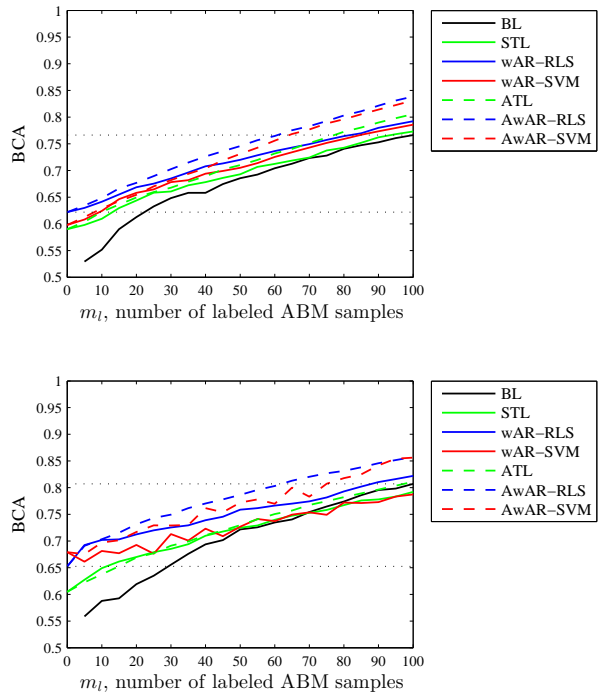}
\caption{Average BCAs of AwAR-RLS and AwAR-SVM for different feature sets, when switching from BioSemi to ABM. Top: 20 nonlinear PCA features; Bottom: 18 theta and alpha band power spectral density features.} \label{fig:features}
\end{figure}

Two other feature sets were employed to study the robustness of AwAR-RLS and AwAR-SVM to different feature extraction methods: 1) 20 nonlinear PCA features extracted from an auto-encoder \cite{Bengio2009}; and, 2) 18 power spectral density features [theta band (4-7.5Hz) and alpha band (7.5-12Hz)] from the 9 common channels using Welch's method \cite{Welch1967}. The BCA results are shown in Fig.~\ref{fig:features}. Observe that AwAR-RLS and AwAR-SVM still achieved the best overall BCAs in both cases, and they had more obvious performance improvements over other methods than the linear PCA case in Fig.~\ref{fig:B2Aavg}. The BCAs of ATL decreased on these two feature sets, suggesting that ATL is not as robust as AwAR-RLS and AwAR-SVM to different features.

\begin{figure}[htpb]\centering
\subfigure[]{\label{fig:sigma}\includegraphics[width=\linewidth]{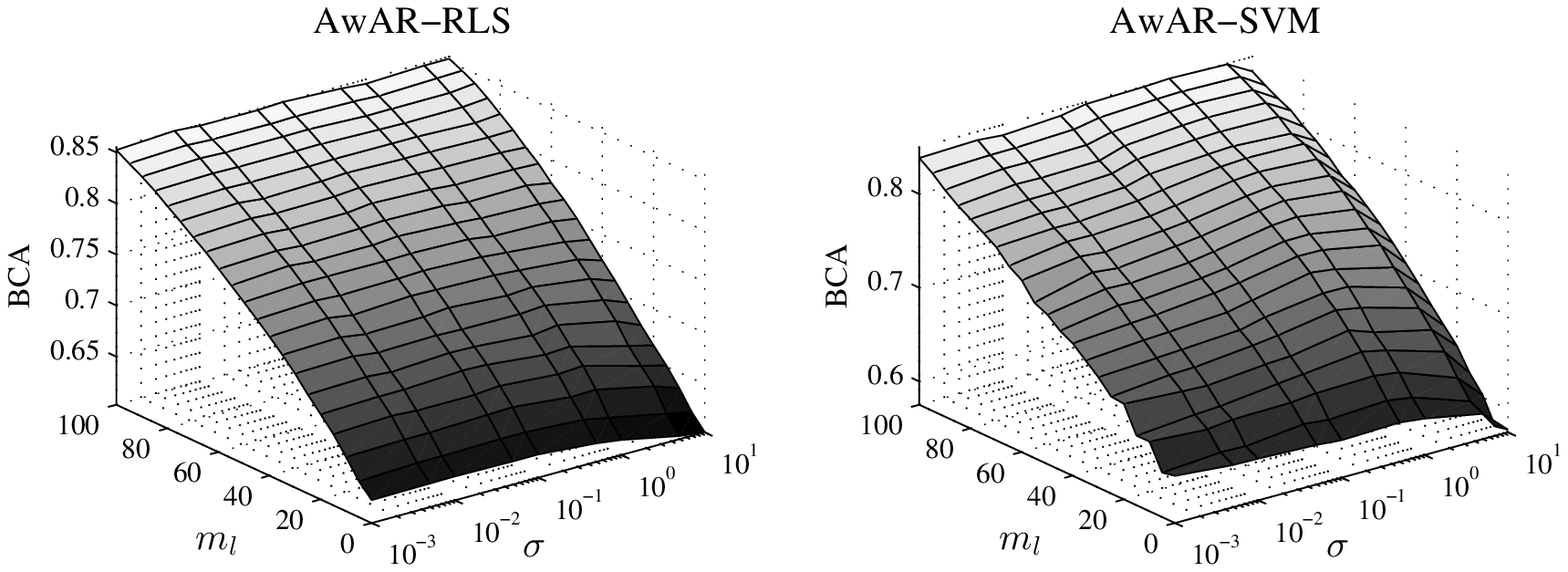}}
\subfigure[]{\label{fig:lambda}\includegraphics[width=\linewidth]{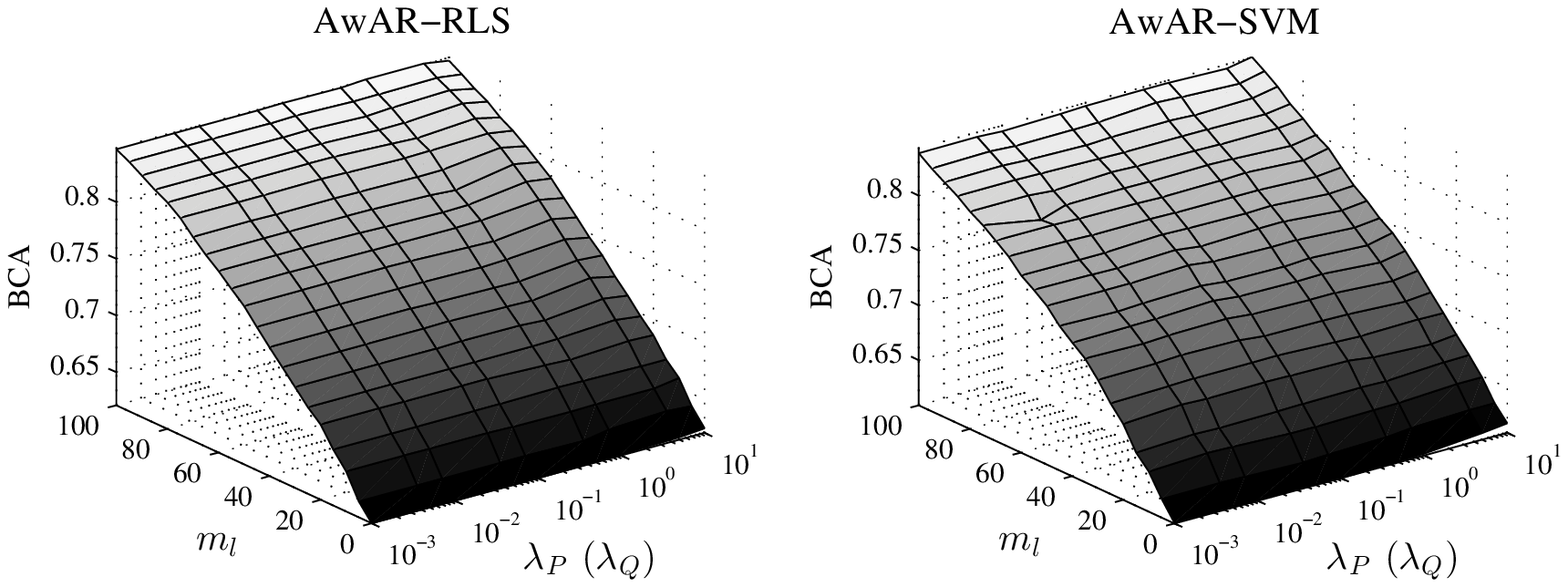}}
\caption{Average BCAs of AwAR-RLS and AwAR-SVM for different parameters, when switching from BioSemi to ABM. (a) $\sigma$; and, (b) $\lambda_P$ and $\lambda_Q$.} \label{fig:parameters}
\end{figure}

The average BCAs of AwAR-RLS and AwAR-SVM for different $\sigma$ ($\lambda_P$ and  $\lambda_Q$ were fixed at 10) are shown in Fig.~\ref{fig:sigma}, and for different $\lambda_P$ and\footnote{We always assigned $\lambda_P$ and $\lambda_Q$ identical value because they are conceptually close.} $\lambda_Q$ ($\sigma$ was fixed at 0.1) are shown in Fig.~\ref{fig:lambda}.  Observe from Fig.~\ref{fig:parameters} that AwAR-RLS and AwAR-SVM are robust to both $\sigma$ and $\lambda_P$ ($\lambda_Q$).

\subsection{Discussions}

Extensive experimental results have demonstrated that AwAR-RLS and AwAR-SVM can indeed reduce the calibration effort when switching to a new EEG headset, and they are very robust. However, they still have some limitations, which will be considered in our future research:
\begin{enumerate}
\item AwAR-RLS and AwAR-SVM assume that the old and new headsets have enough common channels. We will need to quantify the minimum number of common channels for them to work well, and develop approaches to perform transfer for headsets with none or very few common channels, e.g., more sophisticated feature extraction methods that allow compensation from close-by electrodes.
\item In the current study each subject performed the same task in three sessions on three different days, with the subject wearing a different headset each day. The headset difference was the most challenging problem in this transfer learning setting, but there could also be session transfer effects, e.g., nonstationarity of the brain, mind wandering, distraction, human-system mutual adaptation, environment impacts, physical condition changes, electrode re-positioning, etc. In future research we will conduct additional experiments, in which each subject wears the same headset in multiple sessions. By comparing the transfer learning performance between sessions with the same headset and between sessions with different headsets, we can separately study the effects of headset transfer and session transfer.
\end{enumerate}

\section{Conclusions} \label{sect:conclusions}

In this paper, we have introduced two active weighted adaptation regularization approaches, which integrate domain adaptation transfer learning and active learning, to expedite the calibration process when a subject switches to a new EEG headset. Domain adaptation makes use of labeled data from the subject's previous headset, whereas active learning selects the most informative samples from the new headset to be labeled. Experiments on single-trial classification of ERPs using three different EEG headsets showed that active weighted adaptation regularization can significantly improve the classification performance, given the same number of labeled samples from the new headset; or, equivalently, it can effectively reduce the number of labeled samples from the new headset, given a desired classification accuracy.

While the current examples are based on intra-subject transfer (e.g., same-subject, different headsets), our ultimate goal is the application of this approach to more sophisticated preprocessing and feature extraction techniques, such as active weighted adaptation regularization from multiple sources (e.g., use data from other subjects and multiple headsets in a new headset calibration), and the generalization of weighted adaptation regularization to online BCI calibration. Together, these will open the door for a host of applications facilitating BCI technology across a wide range of domains. For example, cross-headset transfer learning, as shown here, will allow data acquired from one research group to be utilized by others, enabling a vast wealth of resources for generating calibration data. To date, this has not been a possible practice due to a wide variety of hardware used in research settings. However, the techniques discussed here not only suggest feasibility, but also lay the foundation for understanding the most critical features of data acquisition hardware which affect transfer and classifier performance. This information can, in turn, be used to further refine and propel the system design industry.

\section*{Acknowledgement}

The authors would like to thank Scott Kerick, Jean Vettel and Anthony Ries at the US Army Research Laboratory (ARL) for designing the experiment and collecting the data.

%\bibliographystyle{IEEETranS} \bibliography{drwubib}
% Generated by IEEEtranS.bst, version: 1.13 (2008/09/30)

\end{document}